\documentclass{article}
\usepackage{spconf,amsmath,graphicx,hyperref}
\usepackage{booktabs}
\usepackage{xcolor}
\usepackage{tipa}
\usepackage{etoolbox}

\makeatletter
\def\name#1{\gdef\@name{\em #1\\}}
\makeatother

\apptocmd{\thebibliography}{\small}{}{}

\title{Tagarela - A Portuguese speech dataset from podcasts}
\name{Frederico Santos de Oliveira$^{\star}$ \qquad Lucas Rafael Stefanel Gris$^{\dagger}$ \qquad Alef Iury Siqueira Ferreira$^{\dagger}$ \\
      Augusto Seben da Rosa$^{\ddagger}$ \qquad Alexandre Costa Ferro Filho$^{\dagger}$ \qquad Edresson Casanova$^{\mathsection}$ \\
      Christopher Dane Shulby$^{\text{\P}}$ \qquad Rafael Teixeira Sousa$^{\star}$ \qquad Diogo Fernandes Costa Silva$^{\dagger}$ \\
      Anderson da Silva Soares$^{\dagger}$ \qquad Arlindo Rodrigues Galvão Filho$^{\dagger}$}

\address{$^{\star}$ Federal University of Mato Grosso (UFMT) 
         $^{\dagger}$ Federal University of Goias (UFG) \\
         $^{\ddagger}$ Paulista State University (UNESP) 
         $^{\mathsection}$ NVIDIA 
         $^{\text{\P}}$ Elsa Speak}
\begin{document}
%
\maketitle
\begin{abstract}
Despite significant advances in speech processing, Portuguese remains under-resourced due to the scarcity of public, large-scale, and high-quality datasets. To address this gap, we present a new dataset, named TAGARELA, composed of over 8,972 hours of podcast audio, specifically curated for training automatic speech recognition (ASR) and text-to-speech (TTS) models. Notably, its scale rivals English’s GigaSpeech (10kh), enabling state-of-the-art Portuguese models. To ensure data quality, the corpus was subjected to an audio pre-processing pipeline and subsequently transcribed using a mixed strategy: we applied ASR models that were previously trained on high-fidelity transcriptions generated by proprietary APIs, ensuring a high level of initial accuracy. Finally, to validate the effectiveness of this new resource, we present ASR and TTS models trained exclusively on our dataset and evaluate their performance, demonstrating its potential to drive the development of more robust and natural speech technologies for Portuguese. The dataset is released publicly\footnote{Available at \href{https://freds0.github.io/TAGARELA/}{https://freds0.github.io/TAGARELA/}} to foster the development of robust speech technologies.

\end{abstract}
\section{Introduction}
\label{sec:intro}

\begin{figure*}[htbp]
    \centering
    \includegraphics[width=0.75\textwidth]{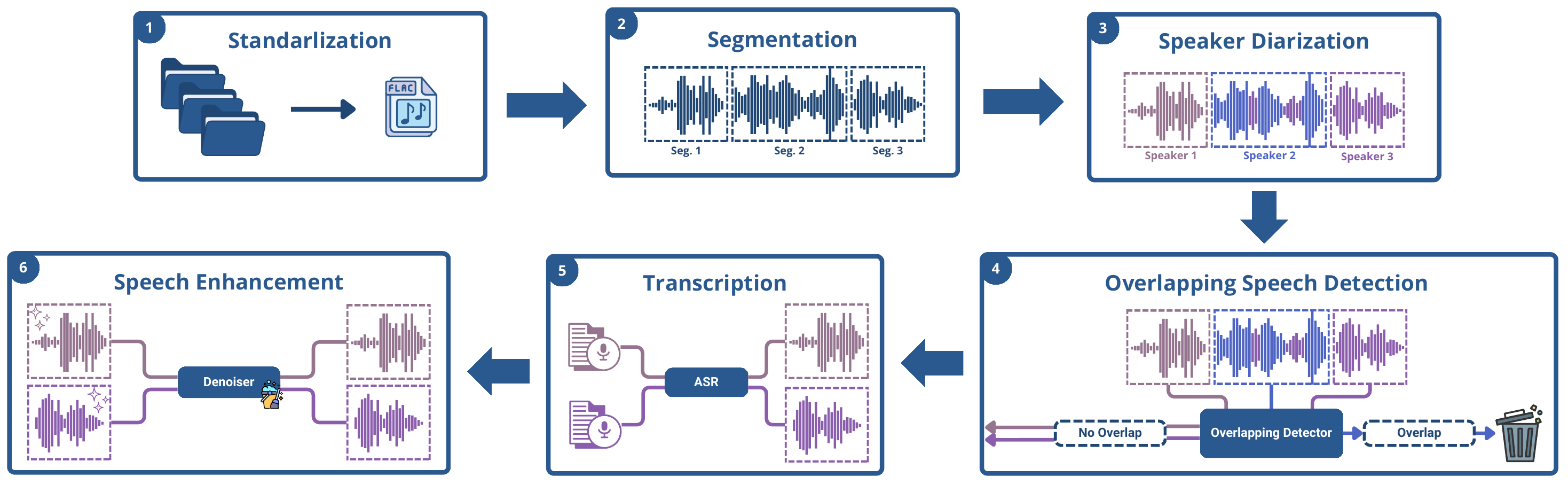}
    \caption{Overview of the TAGARELA preprocessing pipeline.}
    \label{fig:tagarela_pipeline}
\end{figure*}


Portuguese ranks among the most widely spoken languages globally, with hundreds of millions of speakers across several continents. Despite this global prominence, it remains significantly under-resourced in the field of speech technology when compared to English. Recent advances in deep learning have propelled the fields of Automatic Speech Recognition (ASR) and Text-to-Speech (TTS), but their progress is fundamentally driven by the availability of large-scale, high-quality speech datasets. High-resource languages like English benefit from extensive corpora such as LibriSpeech (1000 hours)~\cite{7178964}, GigaSpeech (10k hours)~\cite{10832365} and Emilia (139k hours)~\cite{Chen2021}, which comprise tens of thousands of hours of transcribed audio. This resource gap creates a significant bottleneck that hinders the development of robust and natural-sounding speech technologies tailored to the linguistic nuances of Portuguese.


To address this disparity, the community has released Portuguese corpora focused on spontaneous speech. Landmark initiatives include CORAA~\cite{CandidoJunior2023} (290h), NURC-SP~\cite{nurc-sp-audio-corpus-2024} (239h), MuPe~\cite{evaldo-leal-etal-2025-mupe} (365h), and VoxCeleb-PT~\cite{mendoncca2022voxceleb} (18h). Although these datasets provide high-quality data crucial for ASR, their spontaneous nature — often characterized by disfluencies, background noise, and interruptions — makes them ill-suited for training TTS models, which conventionally require clean, high-quality audio to achieve natural synthesis. Furthermore, being orders of magnitude smaller than their English counterparts, these corpora limit the performance of data-hungry state-of-the-art models.

A significant opportunity to bridge this data gap emerged with the release of the ``Cem Mil Podcasts'' collection~\cite{10.1007/978-3-031-42448-9_5}, a massive corpus offering over 76,000 hours of diverse, multi-dialect Portuguese audio. However, the dataset was provided as raw, unprocessed audio with automatically generated transcripts of varying quality. The absence of essential pre-processing — speaker diarization, noise reduction, and overlapping speech removal — rendered it impractical for training high-performance ASR and TTS models, which require clean, single-speaker segments.

To unlock this resource, we introduce TAGARELA (\textipa{/ta.ga."\textfishhookr E.lA/}), a large-scale Portuguese speech dataset curated from  ``Cem Mil Podcasts''. We developed a pipeline — including standardization, diarization, overlap detection, and denoising — to transform raw audio into a high-quality corpus. For transcription, we employed a bootstrap approach: a 1,000-hour seed corpus, transcribed by commercial ASR, was used to fine-tune Whisper large-v3~\cite{radford2023robust}, generating pseudo-labels for the remaining data. The corpus comprises two subsets: a full 8,972-hour set with disfluencies for robust ASR, and a 2,800-hour clean-speech subset for speech generation.


The main contributions of this work are: (1) the release of TAGARELA, a curated Portuguese speech corpus exceeding 8,972 hours for ASR and TTS; (2) a detailed description and evaluation of our multi-stage processing pipeline; and (3) training and evaluation of open-source models exclusively on TAGARELA, demonstrating its effectiveness.

\vspace{-0.2cm}

\section{Tagarela Dataset}
\label{sec:format}

\vspace{-0.05cm}

\begin{figure}[htbp]
    \centering
    \includegraphics[width=4cm]{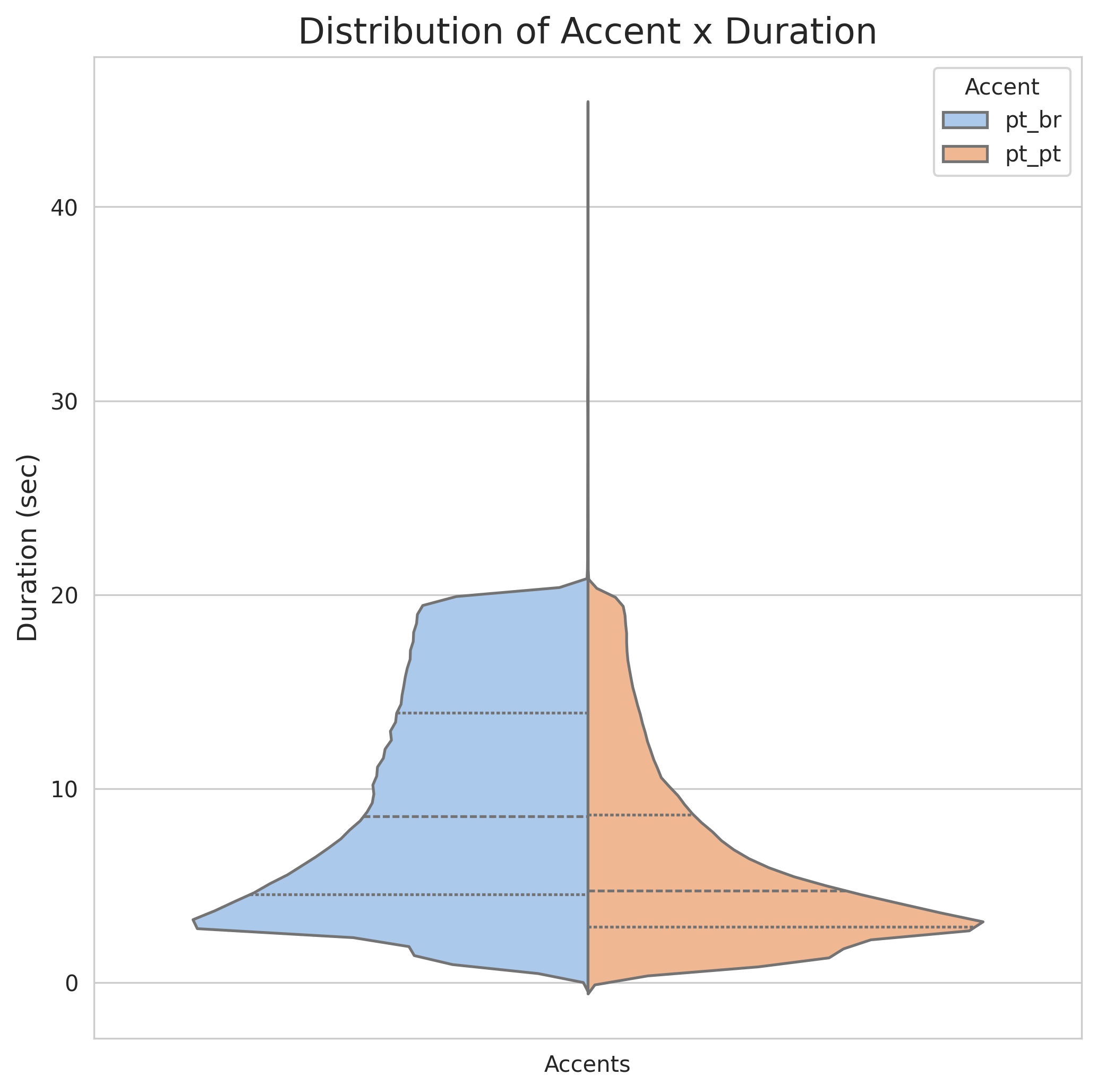}
    \includegraphics[width=4cm]{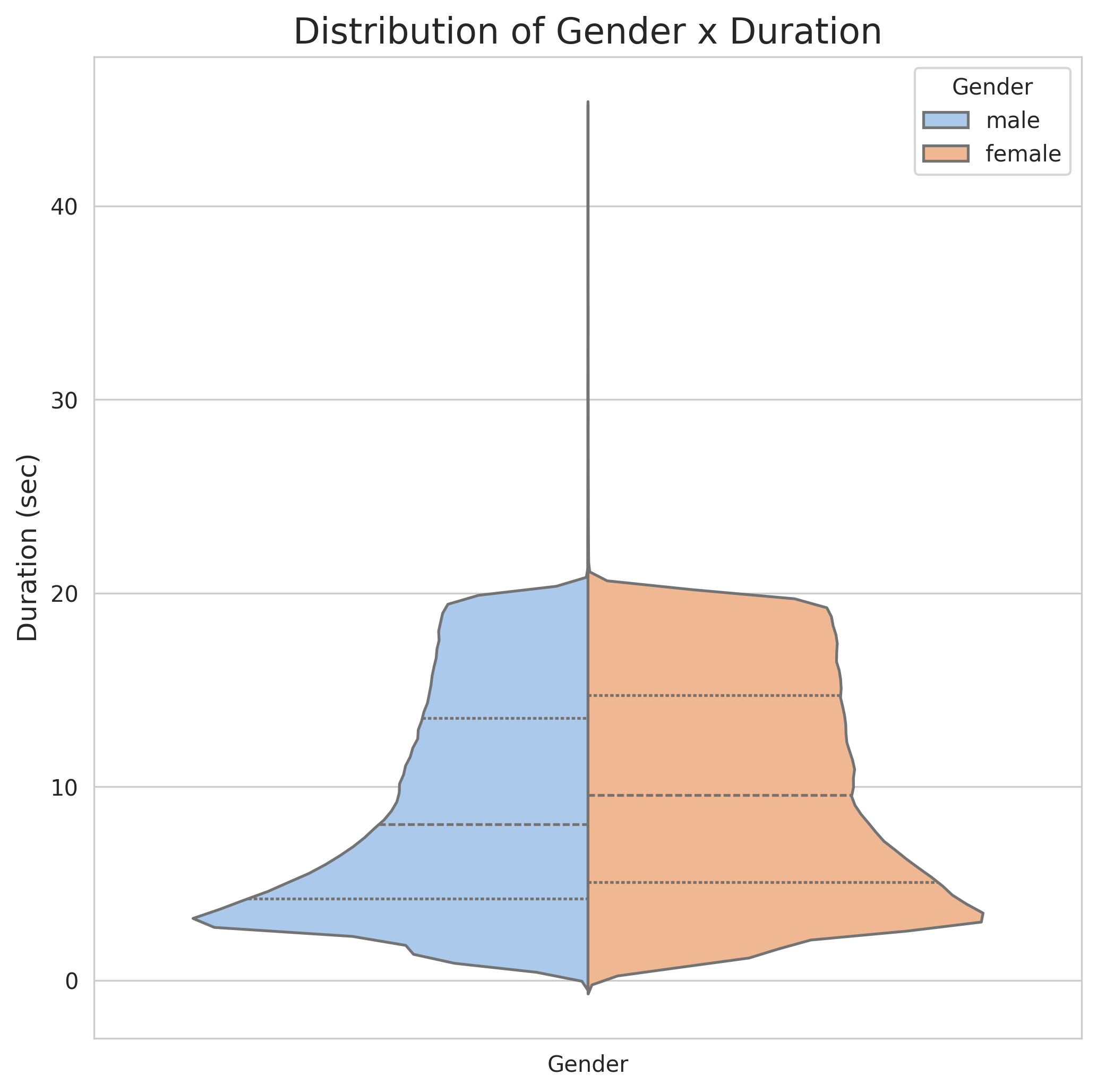}
    \caption{Violin plots showing the distribution of audio segment duration in seconds. The left plot compares accents (pt\_br vs. pt\_pt), and the right plot compares genders (male vs. female).}
    \label{fig:duration_distributions}
\end{figure}

The TAGARELA dataset is a large-scale Portuguese audio corpus derived from the ``Cem Mil Podcasts'' collection \cite{10.1007/978-3-031-42448-9_5} and released exclusively for research purposes to address the lack of non-English podcast resources.
It consists of roughly 16,806 episodes from 2,094 shows, totaling over 8,972 hours of audio. The data includes both Brazilian (8,130 hours) and European Portuguese (842 hours) dialects. In terms of gender, 70\% of the audio (6,368.34 hours) is attributed to male speakers and 30\% (2,604.37 hours) to female speakers. The dataset's audio segments have an average duration of 9.30 $\pm$ 5.49 seconds and contain an average of 27.69 $\pm$ 17.06 words. Figure~\ref{fig:duration_distributions} presents the distribution of audio duration according to gender and accent.

\vspace{-0.5cm}

\section{Tagarela Pipeline}
\label{sec:pagestyle}

The creation of the TAGARELA dataset involved a multi-stage pipeline designed to ensure high quality and consistency for both ASR and TTS tasks. Each stage was planned to handle the challenges inherent to podcast audio, such as multiple speakers, background noise, and the need for accurate, large-scale transcriptions. An overview of the pipeline is shown in Figure~\ref{fig:tagarela_pipeline}, and we detail each component below. 


\vspace{-0.2cm}

\subsection{Audio Standardization and Segmentation}
\label{sec:typestyle}


All audio files were converted to a uniform format (FLAC, 16kHz, 16-bit, mono) to ensure training consistency. Long-form recordings were then segmented into 5–20 second clips, with the algorithm prioritizing splits at natural silences to preserve speech cohesiveness.

\vspace{-0.2cm}

\subsection{Diarization and Speaker Separation}
\label{sec:diarization}

A common feature in podcasts is the presence of multiple speakers, which poses a challenge for creating clean datasets. To address this, we applied a diarization process using the pyannote framework~\cite{pyannote2020}. This stage identifies and labels the speech segments for each speaker individually. By separating the different speakers, we ensure that each final sample in the dataset contains the voice of only one person. This step is particularly crucial for training TTS models, which require single-speaker data to generate consistent voices.

\vspace{-0.2cm}

\subsection{Overlapping Speech Detection}
\label{sec:overlapping_speech}

Although diarization separates speakers, some segments may still contain overlapping speech where multiple speakers talk simultaneously. This is highly detrimental to the quality of TTS models. To mitigate this issue, we trained a dedicated classification model based on Wav2vec2-XLS-R~\cite{Babu2022XLSR} to specifically identify these instances. All audio samples that were flagged by the model as containing overlapping speech were subsequently discarded from the dataset, ensuring that the final clips consist of clean, single-speaker utterances. To ensure the reproducibility of this filtering step, the trained model and its checkpoint are made publicly available for download.

\vspace{-0.3cm}

\subsection{Two-Stage Bootstrap Transcription}
\label{ssec:subhead}

We employed a bootstrap strategy for transcription. First, a high-fidelity ``seed corpus'' of approximately 1000 hours was generated with a commercial ASR service, ElevenLabs Scribe v1\footnote{\href{https://elevenlabs.io/docs/models\#scribe-v1}{https://elevenlabs.io/docs/models\#scribe-v1}, run in June 2025.}, to fine-tune a Whisper large-v3 model for pseudo-labeling. To filter Whisper's potential hallucinations~\cite{Baranski2025}, we also trained a Wav2vec2-XLS-R model\footnote{\href{https://huggingface.co/facebook/wav2vec2-xls-r-1b}{https://huggingface.co/facebook/wav2vec2-xls-r-1b}}~\cite{Conneau2021Unsupervised} on the same seed data. We then calculated the Word and Character Error Rates (WER/CER) between the outputs of both models, making it possible to select only samples with a high agreement score to ensure the training dataset's quality.

\vspace{-0.3cm}

\subsection{Quality Enhancement}
\label{sssec:quality_enhancement}


For the final audio enhancement stage, we repurposed a Vocos vocoder~\cite{ICLR2024_6db0903e} to act as a denoiser. The model was trained specifically for this task on a private dataset, optimizing it to remove common podcast artifacts like background noise, hiss, and light reverberation. This process significantly improves the clarity and quality of the finalized segments. To ensure the reproducibility of this pipeline, we also provide a version of the denoiser trained exclusively on public datasets.



\vspace{-0.3cm}

\subsection{Speaker and Dialect Labeling}
\label{sssec:speaker_dialect_labeling}

To enrich the dataset, we implemented a multi-stage labeling process. First, given that the original data lacks speaker labels, we performed a cluster-based speaker labeling step. We extracted embeddings for each audio segment using the RedimNet B6 model \cite{yakovlev24_interspeech} and grouped them with the HDBSCAN algorithm \cite{9235263}. This process was performed independently for each podcast to avoid merging speaker identities, resulting in approximately 13,368 distinct speaker labels.

Furthermore, we developed a model to classify each segment's dialect as either Brazilian or European Portuguese. This was achieved by first pre-training a wav2vec-base model
on all segmented audio from our dataset. Subsequently, this model was fine-tuned on a balanced combination of the CORAA, CommonVoice~\cite{ardila-etal-2020-common}, and CML-TTS~\cite{oliveira2023cml} datasets to create the final accent classifier
thus adding valuable dialectal metadata to the corpus.

\vspace{-0.2cm}

\section{Experiments and Evaluation}
\label{sec:print}

In this section, we validate the quality and effectiveness of the TAGARELA dataset by using it to train state-of-the-art models for ASR and TTS. Our goal is to demonstrate that the data curated through our pipeline can produce models with competitive or state-of-the-art performance for Portuguese.

\vspace{-0.3cm}

\subsection{Objective Metrics}

\begin{figure}[htbp]
    \centering
    \includegraphics[width=2.8cm]{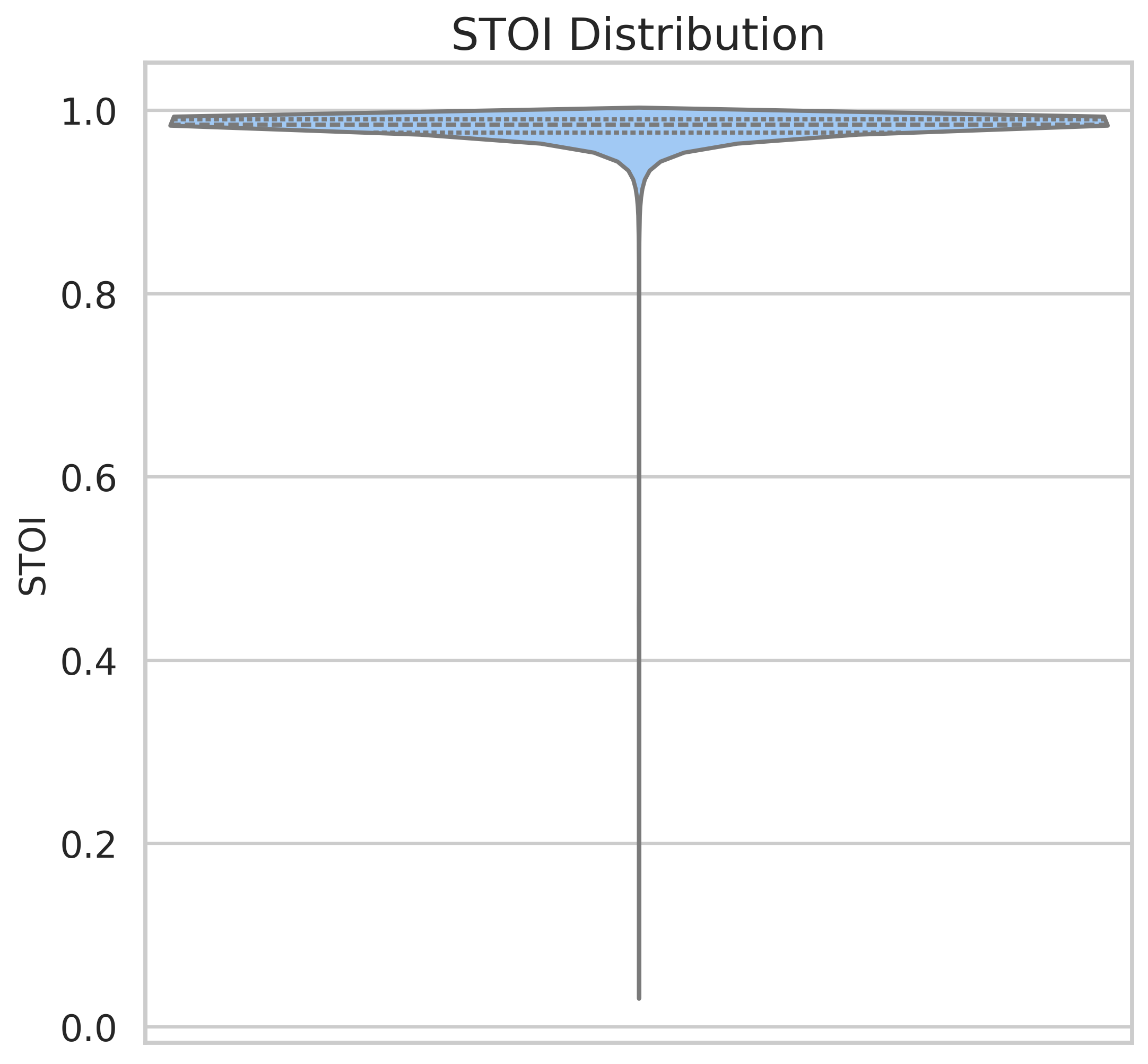}
    \includegraphics[width=2.8cm]{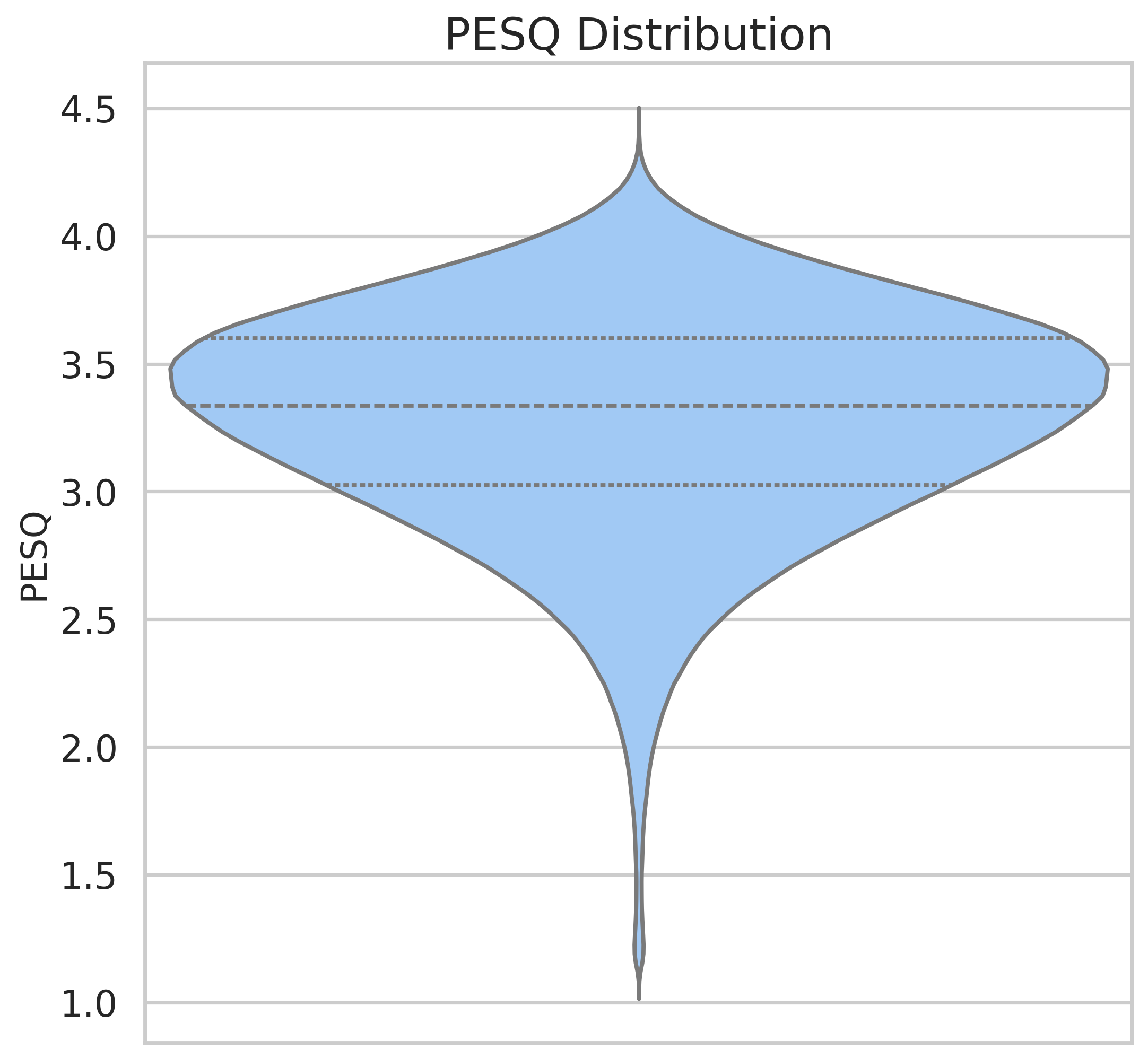}  
    \includegraphics[width=2.8cm]{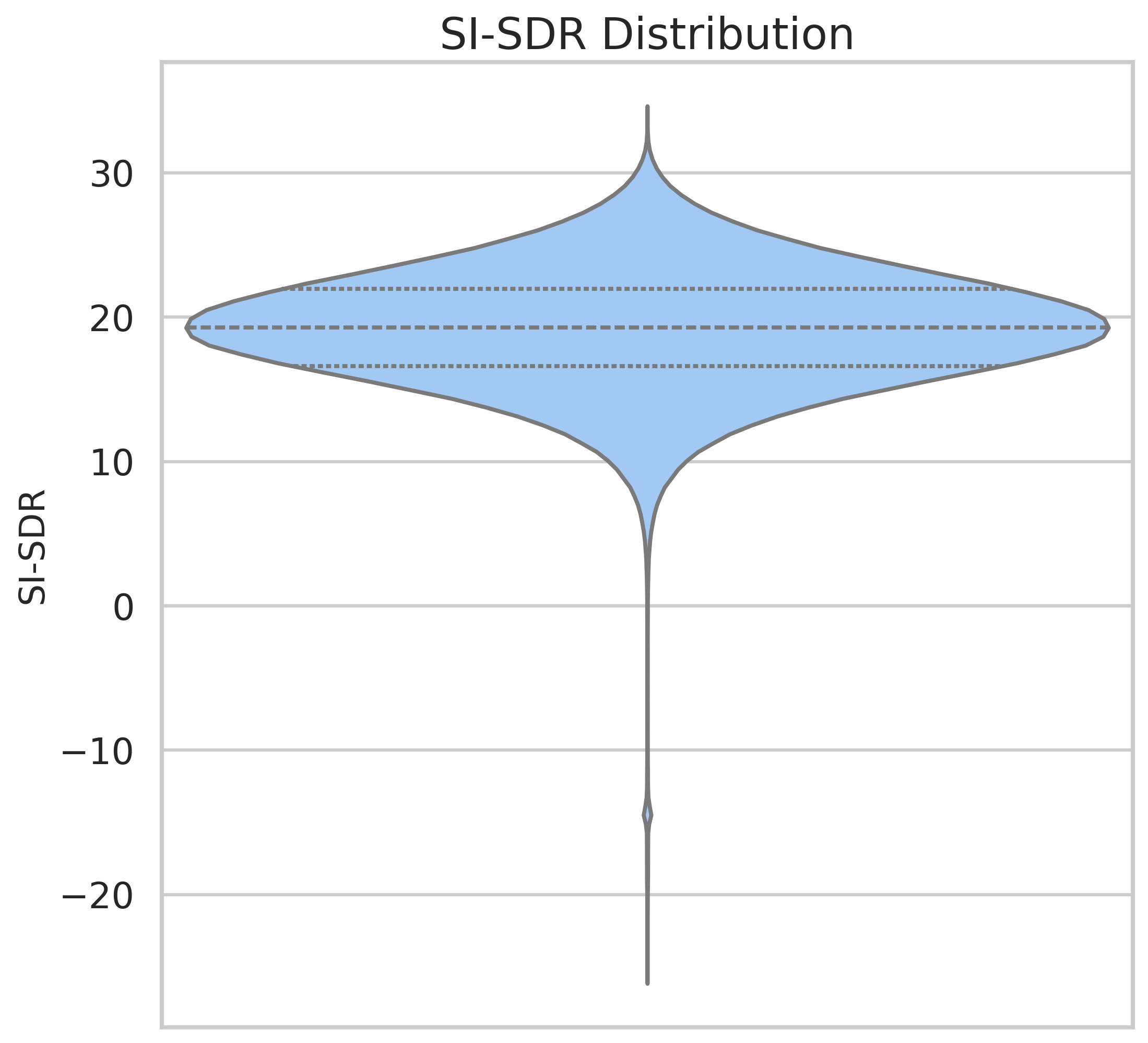}
    \caption{Violin plots showing STOI, PESQ and SI-SDR.}
    \label{fig:squim}
\end{figure}

We assess audio quality using three objective metrics: Short-Time Objective Intelligibility (STOI)~\cite{stoi} for speech intelligibility, the wideband version of Perceptual Evaluation of Speech Quality (PESQ)~\cite{pesq, rec2005wideband} for perceived quality, and Scale-Invariant Signal-to-Distortion Ratio (SI-SDR)~\cite{si_sdr} for signal fidelity in decibels (dB). For all metrics, higher values indicate better quality. Since these traditionally require a clean reference signal, we employ the TorchAudio-Squim~\cite{squim} framework to obtain reference-free estimates. The results are presented in Figure~\ref{fig:squim}.


\vspace{-0.5cm}

\subsection{Speech Recognition (ASR) Experiments}
\label{sec:page}


To evaluate the potential of TAGARELA for ASR tasks, we assessed a diverse set of model architectures covering different sizes and capabilities, including Parakeet TDT~\cite{10389701,10.5555/3618408.3620010}, Wav2Vec Large~\cite{Babu2022XLSR}, Distil-Whisper~\cite{distilwhisper}, Whisper Large V3 \cite{radford2023robust}, and Parakeet v3 \cite{sekoyan2025canary}.

\textit{Training Setup}: We fine-tuned Distil-Whisper, Parakeet TDT v2, and Wav2Vec Large using the full 8,972-hour TAGARELA dataset. In contrast, Whisper Large V3 and Parakeet v3 were evaluated as pre-trained baselines without further training. All experiments were conducted using either NVIDIA A100 or B200 GPUs, subject to availability.

\textit{Evaluation}: We evaluated model performance on the TAGARELA test set manually transcribed, using WER and CER, calculated after text normalization.

\textit{Results}: As detailed in Table~\ref{tab:asr_results}, the finetuned Parakeet v2 yielded the best performance on the TAGARELA test set. Achieving a WER of 15.18\% and CER of 7.09\%, it outperformed all other models, including Wav2Vec Large, Distil-Whisper, and Whisper Large V3, proving to be highly effective for Portuguese speech recognition.


%

\begin{table}[h!]
\centering
\small
\caption{WER results on the TAGARELA test set.}
\label{tab:asr_results}
\begin{tabular}{lcc}
\toprule
\textbf{Model} & \textbf{WER (\%) $\downarrow$} & \textbf{CER (\%) $\downarrow$}  \\
\midrule
Whisper Large V3     & 20.91 & 12.42 \\
Wav2Vec Large FT     & 21.85 & 8.55 \\
Distil-Whisper FT    & 20.02 & 11.18 \\  
Parakeet v3          & 23.30 & 14.86 \\
Parakeet v2 FT       & \textbf{15.18} & \textbf{7.09} \\
\bottomrule
\end{tabular}
\end{table}

\vspace{-0.2cm}

\subsection{Text-to-Speech (TTS) Experiments}
\label{sec:illust}




\textit{Training and Evaluation Setup}: For the TTS task, we trained the Orpheus-TTS\footnote{\href{https://github.com/canopyai/Orpheus-TTS}{https://github.com/canopyai/Orpheus-TTS}} and Chatterbox\footnote{\href{https://github.com/resemble-ai/chatterbox}{https://github.com/resemble-ai/chatterbox}} models using the 2,800-hour clean-speech subset of the TAGARELA dataset. We assessed \textit{intelligibility} using WER/CER (Whisper Large V3) and \textit{perceptual quality} using a Mean Opinion Score (MOS), for which 50 evaluators rated 40 samples. To ensure a robust evaluation, we applied a two-stage outlier removal process, filtering samples shorter than five seconds, and applying a quartile-based method to remove statistical outliers.


\vspace{-0.3cm}
\begin{table}[h!]
\centering
\small
\setlength{\tabcolsep}{4pt} 
\caption{TTS model performance. The values for CER, WER, and MOS are presented as mean $\pm$ standard deviation.}
\label{tab:tts_results}
\begin{tabular}{lccc}
\toprule
\textbf{Model} & \textbf{WER (\%) $\downarrow$} & \textbf{CER (\%) $\downarrow$} & \textbf{MOS $\uparrow$} \\
\midrule
Chatterbox & $0.3111 \pm 0.442$ & $0.268 \pm 0.423$ & $ \textbf{4.176} \pm \textbf{0.983}$\\
Orpheus-TTS & $\textbf{0.095} \pm \textbf{0.100}$ & $\textbf{0.046} \pm \textbf{0.051}$ & $4.155 \pm 1.001$  \\
Ground Truth &  $0.010 \pm 0.033$ & $0.006 \pm  0.018$ & $4.231 \pm 1.001$ \\
\bottomrule
\end{tabular}
\end{table}

\textit{Results and Analysis}: As shown in Table~\ref{tab:tts_results}, Orpheus-TTS achieved superior intelligibility (9.5\% WER), while Chatterbox attained slightly higher naturalness (MOS 4.176) despite significant errors. This reveals a trade-off between Orpheus-TTS's linguistic precision and Chatterbox's focus on prosody.



These experiments validate the TAGARELA dataset's critical role in advancing Portuguese TTS. Despite the dataset's imperfect text-audio alignment, the results are highly encouraging and provide a solid foundation for developing robust, high-quality TTS systems for Portuguese.

\vspace{-0.1cm}

\section{Conclusion}

To address the resource gap in Portuguese speech technology, we introduce TAGARELA, a new large-scale dataset with over 8,972 hours of podcast audio. We presented a comprehensive pipeline using diarization, denoising, and a scalable transcription strategy to create a high-quality corpus suitable for both ASR and TTS. The public release of this dataset is a significant contribution, offering the community a resource on a scale previously unavailable for the Portuguese language.

The effectiveness of TAGARELA was validated by training ASR and TTS models exclusively on our data, achieving highly competitive performance. This confirms the dataset's potential to drive significant advancements in Portuguese speech processing. While there is room for refinements, such as improving text-audio alignment, TAGARELA offers a robust foundation for future innovations. We believe this resource will foster the development of more accurate and natural speech technologies, benefiting millions of Portuguese speakers.

{\bf Acknowledgements:} This work has been fully funded by the project 
Research and Development of Algorithms for Construction of Digital Human Technological Components supported by the Advanced Knowledge Center in Immersive Technologies (AKCIT), with financial resources from the PPI IoT of the MCTI grant number 057/2023, signed with EMBRAPII.

\vspace{-0.3cm}

\bibliographystyle{IEEEbib}
\bibliography{strings,refs}

\end{document}